\documentclass{article} 
\usepackage{iclr2025_conference,times}
\usepackage[utf8]{inputenc} 
\usepackage[T1]{fontenc}    
\usepackage{hyperref}       
\usepackage{url}            
\usepackage{booktabs}       
\usepackage{amsfonts}       
\usepackage{nicefrac}       
\usepackage{microtype}      
\usepackage{xcolor}         
\usepackage{amsfonts}       
\usepackage{amsmath}
\usepackage{amssymb}
\usepackage{amsthm}
\usepackage{amsmath,amssymb,amsthm}
\usepackage{multirow}
\usepackage{graphicx}
\usepackage{natbib}
\usepackage{subcaption}
\usepackage{booktabs}   
\usepackage{makecell}   
\usepackage{adjustbox}
\usepackage{array}    
\usepackage{bbding}
\usepackage{pifont}
\usepackage[toc]{appendix}  
\usepackage{enumitem}
\usepackage{colortbl}
\usepackage{wrapfig}
\usepackage{float}

\usepackage{amsmath,amsfonts,bm}

\def\eqref#1{equation~\ref{#1}}

\def\1{\bm{1}}

\DeclareMathAlphabet{\mathsfit}{\encodingdefault}{\sfdefault}{m}{sl}
\SetMathAlphabet{\mathsfit}{bold}{\encodingdefault}{\sfdefault}{bx}{n}

\usepackage{hyperref}
\usepackage{url}
\iclrfinalcopy
\title{Generalized Few-Shot Out-of-Distribution Detection}

\author{%
  Pinxuan Li, Bing Cao, Changqing Zhang, Qinghua Hu \\
  College of Intelligence and Computing, Tianjin University\\
  Tianjin Key Lab of Machine Learning\\
  \texttt{{pinxuanli, caobing, zhangchangqing, huqinghua}@tju.edu.cn}\\}

\begin{document}
\maketitle

\begin{abstract}
Few-shot Out-of-Distribution (OOD) detection has emerged as a critical research direction in machine learning for practical deployment. Most existing Few-shot OOD detection methods suffer from insufficient generalization capability for the open world. 
Due to the few-shot learning paradigm, the OOD detection ability is often overfit to the limited training data itself, thus degrading the performance on generalized data 
and performing inconsistently across different scenarios.
To address this challenge, we proposed a Generalized Few-shot OOD Detection (\texttt{GOOD}) framework, which empowers the general knowledge of the OOD detection model with an auxiliary General Knowledge Model (GKM), instead of directly learning from few-shot data. 
We proceed to reveal the few-shot OOD detection from a generalization perspective and theoretically derive the \textit{Generality}-\textit{Specificity} balance (\textit{GS}-balance) for OOD detection, which provably reduces the upper bound of generalization error with a general knowledge model.
Accordingly, we propose a Knowledge Dynamic Embedding (KDE) mechanism to adaptively modulate the guidance of general knowledge. KDE dynamically aligns the output distributions of the OOD detection model to the general knowledge model based on the Generalized Belief (G-Belief) of GKM, thereby boosting the \textit{GS}-balance. Experiments on real-world OOD benchmarks demonstrate our superiority. Codes will be available.
\end{abstract}
\section{Introduction}
Deep learning systems are primarily built upon the theoretical framework of the independent and identically distributed assumption, which presumes identical probability distributions between training and test data. However, real-world data acquisition systems inevitably face challenges of distributional shifts, where such discrepancies in probability distributions may pose significant safety risks, particularly in safety-critical applications such as autonomous driving and medical diagnosis. In response to these challenges, diverse methodologies for OOD evaluation have proliferated. Notably, with the advent of prompt learning in pre-trained vision-language models \cite{Radford_Kim_Hallacy_Ramesh_Goh_Agarwal_Sastry_Amanda_Mishkin_Clark_et}, CLIP-based prompt tuning \cite{zhou2022learning, zhou2022conditional} has been strategically adapted for OOD detection tasks, catalyzing growing research interest in leveraging prompt learning paradigms for enhanced OOD detection capabilities.


Recent advances in few-shot OOD detection leveraging Contrastive Language-Image Pre-training  have demonstrated significant progress. Novel few-shot learning methodologies \cite{miyai2024locoop, bai2024id, nie2024out} effectively utilise limited ID data alongside auxiliary OOD samples to enhance model adaptability. However, our experiments reveal a critical limitation of current few-shot learning paradigms for OOD detection - their constrained generalisation capability. Specifically, while these models perform well on specialised OOD datasets, they exhibit substantial performance degradation when evaluated on multi-domain generalised OOD datasets, as demonstrated quantitatively in Fig. \ref{fig1}. This limitation primarily stems from model overfitting to the restricted training distribution, thereby limiting its capacity to recognise diverse unknown OOD patterns in open-world scenarios. One approach to address this challenge is to enhance the diversity of both ID and OOD training samples. However, due to the limitations of the few-shot learning paradigm on real datasets, obtaining large quantities of sufficiently varied OOD samples is unrealistic. Therefore, we propose using a GKM with excellent generalisation capabilities to guide our model. Inspired by this perspective, we propose a theoretical framework that introduces a novel learning paradigm based on GKM.

\begin{figure}[tp]
  \centering
  \includegraphics[width=1\textwidth]{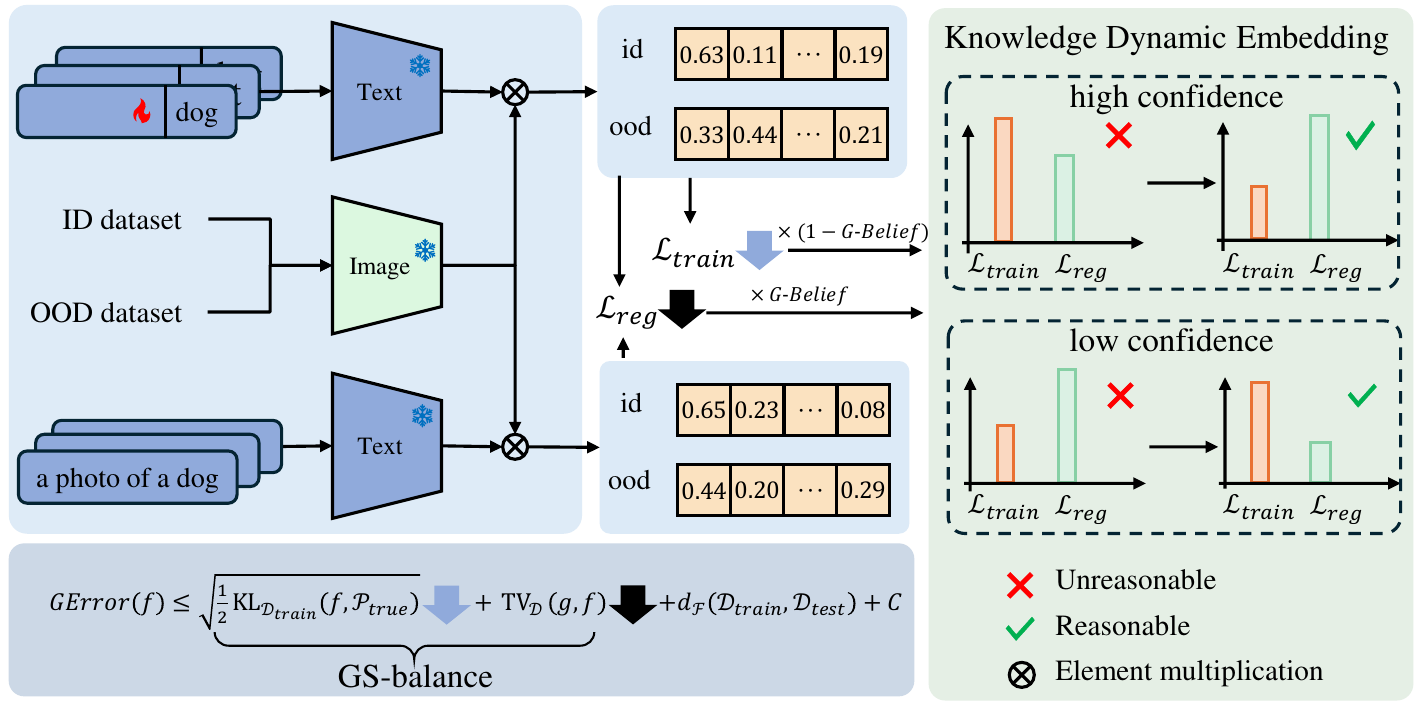}
  \caption{
    Overview of our framework. Our approach initially introduces a distribution regularization loss aligned with general knowledge. Then,  proposed a dynamic credible knowledge learning to explore general knowledge of credible samples, where low-confidence samples allow to explore optimal OOD detection patterns while high-confidence samples preserve reliable general knowledge.}
    \vspace{-3pt}
    \label{fig2}
\end{figure}

In this work, we introduce a Generalized Few-shot OOD Detection
(\texttt{GOOD})framework which is illustrated in Fig. \ref{fig2}. First, we reveal few-shot OOD detection from a generalisation perspective, theoretically deriving the \textit{Generality}-\textit{Specificity} Balance (\textit{GS}-Balance) for OOD detection. This balance is shown to reduce the upper bound of generalisation error with a general knowledge model. Based on this theoretical foundation, we then propose a distribution regularisation loss between the model outputs and the GKM outputs. This regularisation aims to minimise distributional discrepancies between the trained model and the GKM, thereby preserving the latter's general knowledge during few-shot learning. Accordingly, we devise a Knowledge Dynamic Embedding (KDE) mechanism to adaptively modulate the guidance of general knowledge. Specifically, based on the GKM's Generalized Belief (G-Belief), we impose stronger regularisation constraints on samples on which the GKM exhibits high confidence, ensuring the faithful inheritance of reliable general knowledge. Conversely, we relax the regularisation constraints for samples with low confidence, enabling the model to explore optimal OOD detection patterns for uncertain instances. We conduct comprehensive evaluations across diverse OOD benchmarks, including the widely used dataset \cite{Huang_Li_2021}, the OOD dataset setting from OpenOOD \cite{Yang_Wang_Zou_Zhou_Ding_Peng_Wang_Chen_Li_Sun_et}, and the challenging Hard-OOD dataset configured via the MCM protocol \cite{ming2022delving}. Our method demonstrates effective improvements in both conventional and adversarial OOD scenarios. The contributions of this paper are summarised as follows:
\setlength{\belowcaptionskip}{3pt}

\begin{itemize}[parsep=3pt]
    \item This paper provides an intuitive and novel Generalized Few-shot OOD Detection paradigm from the perspective of generalization error. Under theoretical analysis, we derive a new  Generalized Few-shot OOD Detection (\texttt{GOOD}) framework based on the \textit{Generality}-\textit{Specificity} Balance (\textit{GS}-Balance). This offers theoretical guarantees to reduce the upper bound of generalization error in few-shot OOD detection. 
    \item We introduce an auxiliary general knowledge model and propose a Knowledge Dynamic Embedding (KDE) mechanism to adaptively modulate the general knowledge guidance. This offers the generality of the OOD detection model on a specificity basis.
    \item We developed a Generalized Belief (G-Belief) that provides reliable guidance on general knowledge to avoid potential misguidance from the general model. Experiments validated our theoretical analysis and demonstrated the superior performance of our method.
\end{itemize}

%


\section{Related Work}
\label{sec:related}

\textbf{OOD Detection with Pre-trained Vision-language Models.}
Out-of-distribution (OOD) detection aims to identify inputs from unknown classes absent during training, ensuring model reliability. Traditional methods leverage confidence scores like MSP \cite{hendrycks2016baseline}, perturbation-enhanced ODIN \cite{liang2017enhancing}, or feature-space metrics such as Mahalanobis distance \cite{lee2018simple}. Recent advances exploit vision-language models (VLMs) like CLIP, which align visual and textual embeddings for zero-shot inference. Zero-shot CLIP-based approaches \cite{esmaeilpour2022zero}utilize pre-trained prompts to estimate OOD score with temperature-scaled softmax to enhance separability without fine-tuning. Beyond zero-shot, fine-tuned methods \cite{du2022vos}, \cite{tao2023non} incorporate ID data for task-specific calibration, albeit with increased computational costs, such as CLIPN \cite{wang2023clipn} refine detection via negative prompt generation. Recently, a promising direction is few-shot OOD detection, which balances efficiency and performance by leveraging minimal in-distribution samples. Currently, the mainstream approaches leveraging CLIP-based prompt learning for OOD few-shot detection primarily follow two methodologies. One is the LoCoOP \cite{miyai2024locoop} method, which enforces entropy uniformity for distribution alignment, and the other adopts a K+1 class formulation that introduces an auxiliary dimension to learn negative prompts. Variants of the latter include techniques such as id-like \cite{bai2024id}, which reduces the number of negative prompts by learning common features across categories, and the NegPrompt \cite{li2024learning} method, which employs shared class-specific contexts for both positive and negative prompt construction.

\textbf{Prior knowledge transfer.}
Prior knowledge transfer has been extensively utilized across various domains. In the context of CLIP, prior knowledge transfer has been widely adopted to address catastrophic forgetting in tasks such as few-shot accuracy prediction and domain generalization. For instance, ProGrad \cite{zhu2023prompt} ensures alignment between the learning direction of trainable task-specific knowledge and general knowledge (hand-crafted prompts) during prompt tuning, thereby preserving existing knowledge while acquiring new capabilities. Similarly, while ProGrad discards conflicting updates by optimizing prompts toward aligned directions, KgCoOp \cite{yao2023visual} avoids knowledge discardment by introducing a Euclidean distance-based loss to constrain trainable task-specific knowledge to remain proximal to general knowledge. Inspired by these approaches, our work investigates catastrophic forgetting in OOD detection and explores how to effectively leverage prior knowledge transfer to enhance OOD detection performance.

\section{Preliminaries}
\label{sec:Preliminaries}
\newcommand{\Dtrain}{$\mathcal{D}_{\text{train}}$}
\newcommand{\Dtrainid}{$\mathcal{D}^{id}_{\text{train}}$}
\newcommand{\Dtrainood}{$\mathcal{D}^{ood}_{\text{train}}$}
\newcommand{\Dtest}{$\mathcal{D}_{\text{test}}$}
\newcommand{\Dtestid}{$\mathcal{D}^{id}_{\text{test}}$}
\newcommand{\Dtestood}{$\mathcal{D}^{ood}_{\text{test}}$}

We partition the dataset into a training set $\mathcal{D}_{\text{train}} = (\mathcal{D}_{\text{train}}^{\text{ID}}, \mathcal{D}_{\text{train}}^{\text{OOD}})$ and a validation set $\mathcal{D}_{\text{test}} = (\mathcal{D}_{\text{test}}^{\text{ID}}, \mathcal{D}_{\text{test}}^{\text{OOD}})$, where the ID components $\mathcal{D}_{\text{train}}^{\text{ID}}$ and $\mathcal{D}_{\text{test}}^{\text{ID}}$ adhere to the joint data-label distribution $(x_{\text{i}}, y_{\text{i}})$ with explicit sample-label pairs $(x,y)$, while the out-of-distribution (OOD) components $\mathcal{D}_{\text{train}}^{\text{OOD}}$ and $\mathcal{D}_{\text{test}}^{\text{OOD}}$ are sampled from unknown $P_{\text{OOD}}$. But current few-shot learning paradigms increasingly avoid reliance on external OOD datasets. Methods exemplified by LoCoOP \cite{miyai2024locoop} leverage CLIP's inherent prior knowledge to synthesize OOD samples from ID data through patch-based strategies, while approaches like id-like \cite{bai2024id} generate OOD representations via random cropping of ID samples. Consequently, the majority of OOD data in few-shot scenarios originates from systematic transformations of ID data rather than external collections.

\textbf{Zero-Shot OOD Detection.}
 Given a pre-trained vision-language model \cite{Radford_Kim_Hallacy_Ramesh_Goh_Agarwal_Sastry_Amanda_Mishkin_Clark_et} with image encoder $\phi_I(\cdot)$ and text encoder $\phi_T(\cdot)$. The MCM \cite{ming2022delving} method computes OOD scores through cross-modal alignment. MCM's zero-shot capability stems from leveraging the pre-trained cross-modal alignment without fine-tuning on ID data. The key hypothesis is that OOD samples exhibit lower maximum similarity due to semantic misalignment with ID class prompts. For an input image $\mathbf{x}$, the scoring function $S(\mathbf{x})$ is defined as:

\begin{equation}
    S(\mathbf{x}) = \max_{i} \frac{\exp\left(\langle \phi_I(\mathbf{x}), \phi_T(\mathbf{t}_i) \rangle / \tau \right)}{\sum_{j=1}^{C} \exp\left(\langle \phi_I(\mathbf{x}), \phi_T(\mathbf{t}_{j}) \rangle / \tau \right)}
\end{equation}

where $\mathbf{t}_i$ represents the hand-crafted prompt for class $i$, and $\tau$ is the temperature parameter to be set as 1. The OOD decision rule follows:
\begin{equation}
    \mathcal{F}(\mathbf{x}) = 
    \begin{cases} 
        \text{ID}, & S(\mathbf{x}) \geq \tau \\
        \text{OOD}, & S(\mathbf{x}) < \tau 
    \end{cases}
\end{equation}
\textbf{Prompt learning for OOD detection.}
In contrast to conventional prompt learning frameworks like CoOP \cite{zhou2022learning}, our method adopts the state-of-the-art OOD detection approach LoCoOp \cite{miyai2024locoop} . This framework operates without introducing auxiliary dimensions, instead directly fine-tuning the original classification logit distribution. To address the challenge of detecting real OOD samples under distributional uncertainty, we construct auxiliary OOD data by leveraging low-similarity patches from ID samples under the LoCoOp \cite{miyai2024locoop} paradigm. A uniform label distribution $U$ is imposed to suppress the original distribution of OOD data. We use C represents the number of ID classes. Our final loss function is formulated as:
\begin{equation}
\mathcal{L}_{train} = \mathcal{L}_{\text{CE}} +\lambda\mathcal{L}_{\text{OOD}} = \mathbb{E}_{ \mathcal{D}_{\text{train}}^{\text{ID}}}  \Bigg[ -y_{true} \log f_{true}(x;\theta)  \Bigg]+\alpha\mathbb{E}_{\mathcal{D}_{\text{train}}^{\text{OOD}}} \left[ \sum_{i=1}^{C} \frac{1}{C} \log f_i(x;\theta) \right]
\end{equation}

\noindent Here $f_i(x)$ denotes the predicted probability for class $i$, $\lambda$ controls the regularization strength, and $C$ is the number of ID classes.




\section{Generalized Few-shot OOD Detection}
\subsection{Theoretical Foundations of \textit{Generality}-\textit{Specificity} balance}
\newtheorem{theorem}{Theorem}
\newcommand{\pCOV}{p^{\mathrm{COV}}}
\newcommand{\pSEM}{p^{\mathrm{SEM}}}
\newcommand{\GError}{\mathit{GError}}
\newcommand{\ddF}{d_{\mathcal{F}}}
\newcommand{\Lreg}{\mathcal{L}_{\mathrm{reg}}}
\begin{figure}[tp]
  \centering
  \includegraphics[width=0.99\textwidth]{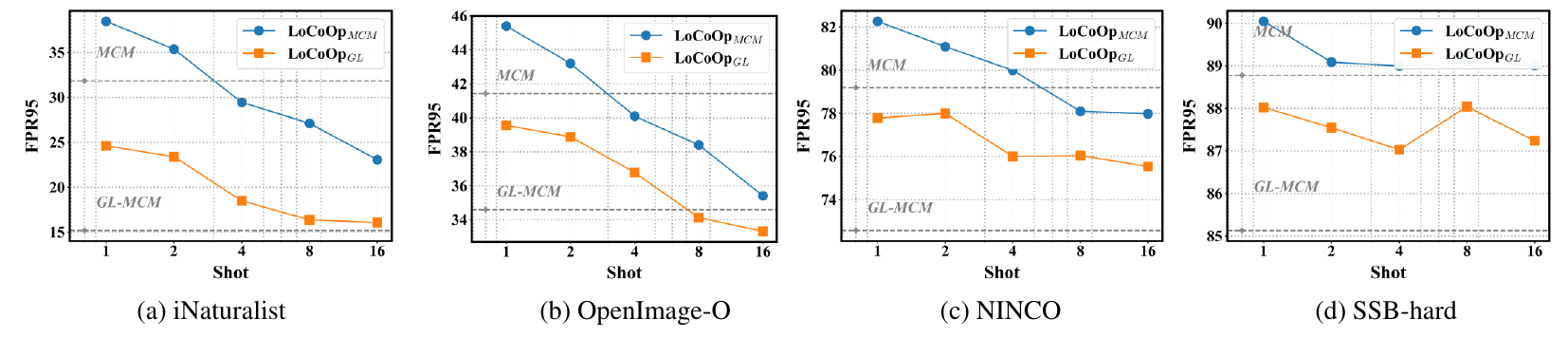}
  \caption{
    Comparative evaluation across diverse OOD benchmarks, including iNaturalist, OpenOOD-proposed Near-OOD, and MCM-configured Hard-OOD datasets. Our method demonstrates state-of-the-art detection performance, outperforming existing approaches in both few-shot adaptability and boundary case robustness.
    }
    \label{fig1}
    \vspace {-2.0em}
\end{figure}

\textbf{Base Setting.}
Given a dataset $\mathcal{D}$ comprising samples $(x, y)$, where $x$ includes both ID samples $x_{\text{id}}$ and auxiliary OOD samples $x_{\text{ood}}$. The ground-truth distribution $\mathcal{P}_{\text{true}}$ corresponds to the true labels for $x_{\text{id}}$ or a uniform distribution $\mathcal{U}$ over $C$ classes for $x_{\text{ood}}$.

To formalize the discrepancy between hypotheses, we define the total variation distance (TV) between two hypotheses $f, f' \in \mathcal{F}$ under distribution $\mathcal{P}$:
\begin{equation}
\label{eq:tvd}
\mathrm{TV}_{\mathcal{D}}(f, f') = \frac{1}{2} \mathbb{E}_{\mathcal{D}} \left[ \sum_{i=1}^C \left| f_i(x) - f'_i(x) \right| \right],
\end{equation}
where $f_i(x)$ and $f'_i(x)$ denote the predicted probabilities for class $i$ by hypotheses $f$ and $f'$, respectively.

Following Yao \cite{yao2024out}, the generalization error of an OOD detector $f$ is defined as:
\begin{equation}
\label{eq:gerror}
\mathrm{GError}(f) = \mathrm{TV}_{\mathcal{D}}(f, \mathcal{P}_{\text{true}}).
\end{equation}

Note that since our ground-truth distribution $\mathcal{P}_{\text{true}}$ corresponds to the true labels for ID samples $x_{\text{id}}$ or a uniform distribution $\mathcal{U}$ over $C$ classes for OOD samples $x_{\text{ood}}$, the generalization error defined here inherently measures the joint generalization capability of both ID classification and OOD detection.

We now establish our primary theoretical contribution: an upper bound on the joint generalization error for specificity knowledge and generality knowledge learning balance for OOD detection.

\begin{theorem}
\label{theorem1}
(\textit{Generality}-\textit{Specificity} balance) Let g denote a general knowledge model. For any hypothesis $ f \in \mathcal{F} $ and $ \delta \in (0,1) $, We holds with a confidence level of $ 1 - \delta $:  

\begin{equation}
GError(f) \leq 
\underbrace{\sqrt{\frac{1}{2}\mathrm{KL}_{\mathcal{D}_{\text{train}}}(f,\mathcal{P}_{\text{true}})}}_{\text{Train Loss}} + 
\underbrace{\mathrm{TV}_{\mathcal{D}}(g,f)\vphantom{\sqrt{\frac{1}{2}\mathrm{KL}_{\mathcal{D}_{\text{train}}}}}}_{\text{Regularization Loss}} + 
d_{\mathcal{F}}(\mathcal{D}_{\text{train}}, \mathcal{D}_{\text{test}}) +
C
\end{equation}
where  $d_{\mathcal{F}}(\mathcal{D}_{\text{train}}, \mathcal{D}_{\text{test}})= \mathrm{TV}_{\mathcal{D}_{\text{test}}}(g, f^{*})-\mathrm{TV}_{\mathcal{D}_{\text{train}}}(g, f^{*})$ quantifies the distributional discrepancy between training and test datasets as perceived by the general knowledge model, measured via TV divergence. $f^{*}$ is the theoretically optimal model. C is constant depending on Rademacher complexities.
\end{theorem}

The proof is deferred in \ref{sec:A}. The generalization error upper bound in our framework is primarily governed by three components: the KL divergence between the training model and groundtruth, the Total Variation distance characterizing the discrepancy between the training model and GKM, and the TV-based distributional shift between training and test datasets as quantified through the lens of the GKM. Where $d$ is a constant determined by the GKM, we refer to the first two items as \texttt{GS}-balance.

\textbf{Remark1.} The KL divergence between the trained model and the ground-truth distribution, along with the TV distance that characterises the deviation of the trained model from the GKM, are pivotal determinants.  Integrating the GKM improves the performance of both ID classification and OOD detection. Reducing the upper bound of generalisation errors requires reducing both terms through systematic optimisation. In practice, however, the first and second terms generally contradict each other, so we refer to the sum of the losses of these two terms as the \texttt{GS}-balance.

\textbf{Remark2.} Our theoretical analysis reveals that selecting a well-generalized knowledge model fundamentally governs the parameter  $d_{\mathcal{F}}(\mathcal{D}_{\text{train}}, \mathcal{D}_{\text{test}})$ in the error bound formulation. Such a model, demonstrating superior cross-domain adaptability, exhibits consistent robustness across diverse OOD datasets. This intrinsic generalization capability enables effective quantification of the distributional shift between training and test environments, consequently maintaining relatively lower magnitudes of $d_{\mathcal{F}}$ even when confronted with challenging hard-OOD scenarios.

\subsection{KDE:Knowledge Dynamic Embedding }
In the preceding section, we established the theoretical foundation for integrating a GKM through derivation of the generalization error upper bound. To minimize the dominant first two terms influencing this bound, we propose the following methodology. Our approach is motivated by  the zero-shot CLIP model inherently exhibits superior OOD detection capabilities, as illustrated in Fig. \ref{fig1}. This insight motivates our selection of the zero-shot CLIP as the GKM, upon which we construct a regularization loss that simultaneously minimizes both the model's training loss and the divergence between its outputs and the knowledge model's logit distributions. To further refine knowledge guidance, we introduce a Knowledge Dynamic Embedding(KDE). Specifically, this mechanism enforces strong alignment with the generalized knowledge on high-confidence samples while adaptively enhancing OOD discriminative power for low-confidence instances. Such  optimization preserves the inherent OOD detection prowess of the zero-shot baseline while enabling progressive adaptation to challenging detection scenarios.

\textbf{Distribution Regularization Loss.}
To address our first objective of preventing excessive logit distribution shifts, we note that the zero-shot model inherently exhibits strong OOD detection capabilities across diverse datasets. Motivated by this observation, we introduce a distribution regularization loss to constrain such shifts. As derived in Theorem \ref{theorem1}, the upper bound of the generalization error depends on the L1 divergence between the fine-tuned model and the high-performing zero-shot baseline. Consequently, we formalize the distribution regularization loss as the L1 divergence between the logits produced by LoCoOp and those generated by the zero-shot model:
\begin{equation*}
\begin{aligned}
\mathcal{L}_{reg}=\frac{1}{C}\sum_{i=1}^{C}|f_{i}-f_{i}^{clip}|
\end{aligned}
\end{equation*}
where $f_{i}$ denotes the logit generated by LoCoOp and ${f}_{i}^{clip}$ represents the logit produced by the zero-shot model.

\textbf{Knowledge Dynamic Embedding.}
However, the direct incorporation of generic knowledge might introduce erroneous information during the training process.   To address this limitation, we propose a dynamic trustworthy learning approach that adaptively leverages reliable knowledge guidance.   For high-confidence samples, we enforce consistency with predictions from the zero-shot model to preserve its robust discriminative capabilities.   Conversely, for low-confidence samples, we mitigate the influence of general knowledge guidance, enabling the model to enhance its OOD detection performance on these uncertain instances. Based on this idea, we first set the Generalized Belief(G-Belief) as:
\begin{equation}
u = \max_{i \in C}f^{\mathrm{clip}}
\end{equation}
Let $u$ denote the zero-shot model’s confidence score for a sample. The overall loss function of our framework is thus formulated as:
\begin{equation}
\mathbb{E}_{(\boldsymbol{x},y)\sim\mathcal{D}_{\text{train}}}\left[{\mathcal{L}_{train}}(f_i(x;\theta)),y)\cdot (1-u)+\lambda\mathcal{L}_{reg}(f_i(x;\theta), f_i^{clip}(x))\cdot u\right]
\end{equation}

\section{Experiments}
\begin{table*}[!t]
\caption{Benchmark OOD detection performance on ImageNet-1K as the ID dataset across CLIP-based architectures. Results are reported as mean across three randomized seeds. ViT-B/16 is adopted as the reference image encoder. }
\label{result_1k_common}
\renewcommand{\arraystretch}{1.05}
\centering
\renewcommand\arraystretch{1.1}
\resizebox{1.0\linewidth}{!}{
\begin{tabular}{cccccccccccc}
\hline
\multirow{3}*{Method} &\multirow{3}*{Backbone} &\multicolumn{10}{c}{OOD Dataset} \\
~ & ~ & \multicolumn{2}{c}{iNaturalist} & \multicolumn{2}{c}{SUN} & \multicolumn{2}{c}{Places} & \multicolumn{2}{c}{Texture} &\multicolumn{2}{c}{Average}\\
\cmidrule(lr){3-4} \cmidrule(lr){5-6} \cmidrule(lr){7-8}  \cmidrule(lr){9-10} \cmidrule(lr){11-12} 
~ & ~           &FPR95	&AUROC  &FPR95	&AUROC  &FPR95	&AUROC  &FPR95	&AUROC  &FPR95	&AUROC\\

\hline
& & \multicolumn{10}{c}{Full/Sub Data Fine-tune}\\
MSP &CLIP-B/16              &$40.89$  &$88.63$  &$65.81$	&$81.24$  &$67.90$  &$80.14$  &$64.96$  &$78.16$  &$57.92$	&$82.31$\\
Energy &CLIP-B/16           &$29.75$  &$94.68$  &$34.28$	&$93.15$  &$56.40$  &$85.60$ &$51.35$  &$88.00$  &$45.83$  &$89.43$\\
ODIN & CLIP-B/16            &$30.22$	&$94.65$	&$54.04$	&$87.17$	&$55.06$	&$85.54$	&$51.67$	&$87.85$	&$45.65$  &$89.35$\\
Fort/MSP & CLIP-B/16        &$54.05$	&$87.43$	&$54.12$	&$86.37$	&$72.98$	&$78.03$	&$68.85$	&$79.06$	&$65.29$  &$81.51$\\
VOS & CLIP-B/16             &$31.65$	&$94.53$   &$43.03$  &$91.92$	&$41.62$	&$90.23$	&$56.67$	&$86.74$	&$43.31$  &$90.50$\\
NPOS & CLIP-B/16            &$16.58$  &$96.19$  &$43.77$	&$90.44$  &$45.27$  &$89.44$  &$46.12$  &$88.80$	&$35.99$  &$91.48$ \\
CLIPN & CLIP-B/16           &$23.94$	&$95.27$ &$26.17$	&$93.93$  &$33.45$	&$92.28$  &$40.83$	&$90.93$  &$32.74$  &$92.83$\\
\hline
& & \multicolumn{10}{c}{Zero-shot}\\
MCM&CLIP-B/16               &$30.91$	&$94.61$	&$37.67$	&$92.56$	&$44.69$  &$89.77$	&$57.77$	&$86.11$	&$44.46$	&$90.16$\\
\hline
& & \multicolumn{10}{c}{One-shot}\\
CoOp & CLIP-B/16            &$43.38$	&$91.26$	&$38.53$  &$91.95$	&$46.68$  &$89.09$	&$50.64$  &$87.83$	&$46.90$	&$89.39$\\
id-like & CLIP-B/16         &$12.07$  &$97.65$  &$40.55$ &$91.07$ &$47.94$ &$88.31$  &$38.34$  &$89.67$  &$34.72$  &$91.67$\\
NegPrompt& CLIP-B/16        &$65.03$ &$84.56$ &$44.39$  &$89.63$  &$51.31$  &$86.55$  &$87.60$  &$63.76$ &$62.08$  &$81.13$\\

LoCoOp$_\texttt{MCM}$ & CLIP-B/16      &$42.15$	&$91.65$	&$31.95$	&$93.74$	&$38.90$  &$90.93$	&$49.05$	&$89.16$	&$40.54$  &$91.62$ \\
LoCoOp$_\texttt{GL}$ & CLIP-B/16        &$28.23$  &$94.51$	&$25.26$  &$94.80$  &$33.63$	&$91.53$	&$51.06$  &$87.39$  &$34.57$	&$91.06$ \\
\rowcolor{gray!40}
\texttt{\texttt{GOOD}}$_\texttt{MCM}$ & CLIP-B/16       &$28.60$	&$94.10$  &$33.44$	&$93.23$	&$41.57$  &$90.26$	&$49.77$  &$88.84$  &$\textbf{38.34}$	&$91.60$\\
\rowcolor{gray!40}
\texttt{GOOD}$_\texttt{GL}$ & CLIP-B/16                 &$15.89$	&$96.30$  &$25.42$	&$94.30$	&$34.43$  &$91.20$	&$49.15$	&$86.28$ &$\textbf{31.22}$	&$\textbf{92.02}$ \\
\hline
& & \multicolumn{10}{c}{16-shot} \\
CoOp & CLIP-B/16            &$35.36$	&$92.60$  &$37.06$	&$92.27$  &$45.38$	&$89.15$  &$43.74$	&$89.68 $  &$41.49$  &$90.48$ \\
id-like & CLIP-B/16         &$13.94$  &$95.42$	&$42.28$	&$89.42$	&$53.25$	&$85.44$	&$18.16$  &$93.78$ &$31.91$  &$91.01$ \\
NegPrompt& CLIP-B/16        &$37.79$  &$90.49$ &$32.11$  &$92.25$  &$35.52$  &$91.16$  &$43.93$  &$88.38$  &$37.34$  &$90.57$ \\
LoCoOp$_\texttt{MCM}$ & CLIP-B/16       &$21.79$  &$95.77$	&$32.40$	&$93.60$  &$41.43$	&$90.57$	&$41.84$	&$91.01$  &$34.36$	&$92.74$ \\
LoCoOp$_\texttt{GL}$ &CLIP-B/16         &$16.88$	&$96.61$  &$23.71$	&$94.93$	&$33.43$  &$91.85$  &$43.54$	&$89.08$  &$29.39$	&$93.12$ \\
\rowcolor{gray!40}
\texttt{GOOD}$_\texttt{MCM}$ & CLIP-B/16        &$26.16$	&$94.38$	&$31.55$	&$93.37$	&$39.01$  &$90.23$  &$39.22$	&$90.99$  &$\textbf{33.99}$	&$92.24$\\
\rowcolor{gray!40}
\texttt{GOOD}$_\texttt{GL}$ & CLIP-B/16         &$17.19$	&$96.09$  &$23.61$	&$94.72$  &$32.77$  &$91.29$  &$41.77$	&$89.41$  &$\textbf{28.83}$  &$92.88$ \\

\hline
\end{tabular}
\vspace {-2.0em}}
\end{table*}

\subsection{Experimental Setup}
\textbf{Datasets.} We evaluate our method across three distinct OOD detection benchmarks to comprehensively assess performance under varying scenarios. First, following standard protocols, we employ ImageNet-1K \cite{deng2009imagenet} as the ID dataset and test on widely-used OOD benchmarks including iNaturalist \cite{van2018inaturalist}, SUN \cite{Xiao_Hays_Ehinger_Oliva_Torralba_2010}, Places \cite{Zhou_Lapedriza_Khosla_Oliva_Torralba_2018}, and Textures \cite{Cimpoi_Maji_Kokkinos_Mohamed_Vedaldi_2013} with few-shot training. Second, to rigorously examine hard OOD detection, we adopt the MCM \cite{esmaeilpour2022zero} ImageNet-10 and ImageNet-20 setup, where ImageNet-10 mimics CIFAR-10’s class distribution with high-resolution images, and ImageNet-20 introduces semantically similar near-OOD classes. Third, for multi-scale OOD analysis, we follow OpenOOD \cite{Yang_Wang_Zou_Zhou_Ding_Peng_Wang_Chen_Li_Sun_et} Setting. We used ImageNet-1K as the ID dataset, SSB-hard \cite{Vaze_Han_Vedaldi_Zisserman_2021}, NINCO \cite{Bitterwolf_Üller_Hein} and OpenImage-O \cite{Wang_Li_Feng_Zhang} as the OOD dataset, evaluating robustness against both subtle semantic shifts and domain discrepancies. All results are averaged over three independent trials to ensure statistical reliability.
\begin{table}[!t] 
\centering
\captionsetup{type=table, skip=4pt}
\caption{cross-domain OOD detection performance comparison across OOD datasets which under different detection frameworks setting: evaluations follow the OpenOOD benchmark with ImageNet-1K as ID data against SSB-hard, NINCO,  and OpenImage-O OOD splits, and the MCM cross-evaluation protocol adopting ImageNet-10 ImageNet-20 as ID datasets with reciprocal OOD testing . Our first row represents the id dataset and the second row represents the ood dataset.}
\label{result_1k_ninco}
\vspace {0.3em}
\setlength{\tabcolsep}{1.4pt} 
\small
\begin{tabular}{@{}lccccccccccccc@{}}
\toprule
\multirow{3}{*}{Method} & \multicolumn{2}{c}{\textbf{ImageNet-10}} & \multicolumn{2}{c}{\textbf{ImageNet-20}} & \multicolumn{6}{c}{\textbf{ImageNet-1K}}  & \multicolumn{2}{c}{\textbf{Average}}\\
\cmidrule(lr){2-3} \cmidrule(lr){4-5} \cmidrule(lr){6-11}
&\multicolumn{2}{c}{\textbf{ImageNet-20}}   &\multicolumn{2}{c}{\textbf{ImageNet-10}} & \multicolumn{2}{c}{\textbf{SSh-hard}} &\multicolumn{2}{c}{\textbf{NINCO}} & 
\multicolumn{2}{c}{\textbf{OpenImage-O}}   \\
\cmidrule(lr){2-3} \cmidrule(lr){4-5} \cmidrule(l){6-7} \cmidrule(l){8-9} \cmidrule(l){10-11} \cmidrule(l){12-13}
& FPR95 & AUROC& FPR95 & AUROC& FPR95 & AUROC & FPR95 & AUROC & FPR95 & AUROC & FPR95 & AUROC \\
\midrule
MCM      & $5.60$  &  $98.86$ & $11.90$  &  $97.68$ & $88.77$  & $64.41$ & $79.24$  & $74.10$ & $41.42$ & $91.84$ & $45.39$  & $85.38$\\
LoCoOp   & $28.20$ & $92.75$ & $34.40$ & $92.34$& $90.27$  & $63.16$ & $82.54$  & $69.19$ & $45.12$  & $90.73$  & $56.11$  & $81.63$ \\
Ours     & $5.70$ & $98.60$ & $16.10$ & $97.66$& $88.78$  & $64.41$ & $79.19$  & $74.10$ & $41.43$  & $91.84$ & $46.24$  & $85.32$\\
\bottomrule
\end{tabular}
\vspace {-1.0em}
\hfill

\end{table}\


\textbf{Implementation details.} Our implementation adheres to the LoCoOp framework with CLIP-ViT-B/16 \cite{Dosovitskiy_Beyer_Kolesnikov_Weissenborn_Zhai_Unterthiner_Dehghani_Minderer_Heigold_Gelly_et} as the backbone, where the feature maps exhibit a spatial resolution of 14×14. The key hyperparameters are empirically configured as follows: the neighborhood size K=200 across all experiments, the knowledge distillation coefficient $\alpha$ =0.25, and the regularization weight $\lambda$ =0.3. Additional training specifications include 50 epochs with a base learning rate of 0.002, batch size of 32, and prompt token length N=16. All experiments are conducted on a single NVIDIA A6000 GPU to ensure hardware consistency.

\textbf{Baselines and Evaluation.} Our comparative analysis encompasses three methodological paradigms. Fully-supervised approaches such as MSP \cite{hendrycks2016baseline}, Fort/MSP \cite{Fort_Ren_Lakshminarayanan_2021}, Energy \cite{Liu_Wang_Owens_Li_2020}, ODIN \cite{liang2017enhancing}, VOS \cite{du2022vos}, and NPOS \cite{tao2023non}, zero-shot approaches represented by MCM \cite{ming2022delving}, and few-shot methods including CoOp \cite{zhou2022learning} and LoCoOp. All methods employ the CLIP ViT-B-16 backbone to ensure equitable comparison. Performance evaluation leverages three standard metrics — FPR95, AUROC, and ID classification accuracy — to enable comprehensive assessment of detection capabilities.

\subsection{Main results}
\textbf{ImageNet-1k as ID dataset.} Table \ref{result_1k_common} summarizes our OOD detection performance using ImageNet-1K as ID data.  Our proposed framework, which synergistically integrates CLIP's intrinsic OOD detection capability with LoCoOp's adaptive mechanism, achieves state-of-the-art performance across both 1-shot and 16-shot configurations.  Particularly noteworthy is the GL-MCM enhanced LoCoOp variant, which demonstrates significant improvements with average FPR95 and AUROC scores of 31.22 and 92.02 in 1-shot settings. This method outperforming conventional  OOD detection methods and even surpassing the original zero-shot CLIP baseline and LoCoOp.  The framework also notably maintains superior ID classification accuracy, confirming that our loss formulation preserves discriminative power for known classes.

\textbf{Other large-scale benchmark OOD dataset.} To validate cross-dataset generalization, we further evaluate on standard OOD benchmarks set by Openood \cite{Yang_Wang_Zou_Zhou_Ding_Peng_Wang_Chen_Li_Sun_et}.  As shown in subsequent Table \ref{result_1k_ninco}, our method consistently matches or exceeds CLIP's native OOD detection performance without dataset-specific tuning.  This empirically validates our key design principle that anchoring adaptation through CLIP's frozen representations successfully transfers its universal OOD awareness to downstream tasks while circumventing catastrophic forgetting of pre-trained knowledge.

\textbf{Comparisons on hard OOD detection. } Our method demonstrates robust performance on small hard-OOD datasetsby maintaining minimal deviation in logit distributions from the zero-shot model, thus preserving its inherent OOD detection capabilities. The result as shown in Table \ref{result_1k_ninco}. Notably, while LoCoOp suffers significant performance degradation on these datasets, our approach consistently matches or even surpasses the zero-shot baseline, achieving superior OOD detection accuracy. This observation confirms that our framework effectively anchors the optimization trajectory near the zero-shot model’s original solution space, avoiding detrimental shifts toward local optima that compromise generalization on specific OOD distributions.

\subsection{Ablation study}
\textbf{Importance of distribution regularization loss.} As the results are shown in the Table \ref{abalation}. As demonstrated in our ablation studies targeting the two core components of our method – the distributional regularization loss and the confidence-based boundary exploration mechanism – we present aggregated experimental results across benchmark datasets.    Our findings reveal that while the proposed distributional regularization loss guarantees superior performance on hard-OOD datasets, its impact remains statistically insignificant on common OOD benchmarks, suggesting this component primarily ensures consistent generalization across diverse OOD test scenarios rather than boosting absolute performance.   

\textbf{Importance of dynamic credible knowledge learning method.} Introducing our dynamic credible knowledge learning strategy after the introduction of distributional regularization loss results in a significant performance improvement on the standard OOD evaluation set while maintaining the benefits of cross-distribution generalization. As the results are shown in the Table \ref{abalation}, our approach enables the model to obtain performance improvement on OOD detection, and this performance improvement stems from the fact that we fully utilize the general knowledge guidance of the high-confidence samples and the self-exploratory nature of the unconfident samples, which effectively coordinates the distributional robustness and the discriminative power of the OOD detection task.

\textbf{Influence of weight of distribution regularization loss.} The hyperparameter $\lambda$  critically governs the distance metric between the text-derived knowledge space and the zero-shot model's latent representation space. To systematically analyze this relationship, we conduct ablation studies with $\lambda$ = [0, 0.1, 0.2, 0.3, 0.4, 0.5], with comprehensive results visualized in Fig. \ref{fig3}. A particularly noteworthy observation emerges from the $\lambda$  = 0.1 configuration, performance degradation occurs due to the induced intermediate embedding space residing in the suboptimal region between the zero-shot prior and LoCoOP's self-learned representations. This phenomenon suggests that weak regularization fails to achieve effective alignment with either the text-guided semantic structure or the discriminative features autonomously discovered by the model, thereby creating an ambiguous optimization landscape that undermines both knowledge sources' complementary strengths.

\begin{table}[!t] 
\begin{minipage}[!t]{0.48\textwidth}
\centering
\captionsetup{type=table, skip=4pt, width=0.85\textwidth}
\caption{Accuracy comparison of ID on the ImageNet-1K validation data for few-shot object detection.}
\label{tab:id_results}
\setlength{\tabcolsep}{12pt} 
\small
\begin{tabular}{@{}lcc@{}}
\toprule
Shot & 
Method & Accuracy (\%) \\
\midrule
\multirow{3}{*}{1-Shot}
&CoOp & 66.23 \\
&LoCoOp & 67.6 \\
&Ours & 66.7 \\
\toprule
\multirow{3}{*}{16-Shot}
&CoOp & 72.10 \\
&LoCoOp & 71.4 \\
&Ours & 70.3 \\
\bottomrule
\end{tabular}
\end{minipage}
\begin{minipage}[!t]{0.48\textwidth}
\centering
\captionsetup{type=table, skip=4pt}
\caption{Ablation analysis of framework components with distribution regularization loss and Knowledge Dynamic Embedding on different OOD datasets with average results.}
\label{abalation}
\setlength{\tabcolsep}{4pt}
\small
\begin{tabular}{@{}lcccccc@{}}
\toprule
 & & & \multicolumn{2}{c}{OOD Benchmark}  & \multicolumn{2}{c}{Other Benchmark} \\
 \cmidrule(lr){4-5}\cmidrule(lr){6-7}
 &$\mathcal{L}_{reg}$ &u & FPR95 & AUROC & FPR95 & AUROC \\
\midrule
&\ding{55} &\ding{55} & $40.54$  & $91.62$  & $56.11$  & $81.63$ \\
&\Checkmark &\ding{55} &$40.33$  & $91.39$  & $45.77$  & $85.46$ \\
&\Checkmark &\Checkmark& $38.34$  & $91.60$  & $46.24$  & $85.32$ \\

\bottomrule
\end{tabular}
\end{minipage}
\vspace {-1.0em}
\end{table}

\subsection{Discussions}
\textbf{The selection of general knowledge model.} We propose to select a model with strong generalization capabilities as the general knowledge model to guide our framework. Specifically, We choose the zero-shot model as the general knowledge model for our few-shot task. As demonstrated in Fig . \ref{fig2} and its superior performance on the different benchmark, the zero-shot baseline outperforms directly trained models like LoCoOp on real-world datasets, exhibiting  OOD detection generalizability. This phenomenon underscores the systemic underestimation of zero-shot model intrinsic OOD detection capacity. We think that the performance degradation in LooCoOp may stem from limited training samples inadvertently distorting the zero-shot model’s original output distribution. While minimizing training loss,  this training method amplifies the total variation distance between the training model and general knowledge model, thereby degrading generalization across diverse OOD datasets. Theoretically, we argue that selecting an appropriate general knowledge model under principled guidance can effectively tighten the upper bound of generalization error. To validate this claim, we conduct extensive ablation studies in \ref{sec:D}, comparing various general knowledge model candidates to verify the effectiveness of our approach.

\textbf{The discussion of reusing a prior knowledge.} 
Current few-shot methods, such as LoCoOp and ID-Like, leverage zero-shot as prior knowledge. However, our distinction lies in the application of prior knowledge: their approaches primarily focus on utilizing such priors to partition ID and OOD datasets through image-level operations. This methodology inevitably introduces distributional shifts in the model's OOD output due to the inherent limitations of finite training data, as exemplified in our experimental results where significant deviation is observed when $\lambda$=0. Consequently, their framework fails to guarantee global optimality on OOD data. In contrast, our method strategically employs prior knowledge in a complementary manner – while existing techniques predominantly operate on the data manifold, our innovation resides in textual space guidance. This orthogonal design principle ensures that our approach remains non-conflicting with LoCoOp's methodology, as they address different aspects of the few-shot learning paradigm.

\textbf{The limitations of the exploration scope.} 
The introduction of distribution regularization in our method inherently bounds the exploration of OOD detection performance within the vicinity of the zero-shot model’s distribution.  While this constraint ensures stability, it may limit the upper detection bounds on specific datasets.  For instance, on the texture dataset, our method—though outperforming the zero-shot baseline—falls short of LoCoOp’s results, as depicted in Table \ref{result_1k_common}. We hypothesize that LoCoOp’s unconstrained exploration in broader parameter spaces could achieve higher OOD detection bounds for such datasets.  However, this potential gain risks degrading performance on other OOD benchmarks, including iNaturalist, a trade-off our framework intentionally avoids.  We posit that sacrificing marginal gains on niche distributions is a necessary compromise to preserve cross-dataset robustness and prevent catastrophic forgetting of the zero-shot model’s inherent capabilities.

\begin{figure}[tp]
  \centering
  \includegraphics[width=0.99\textwidth]{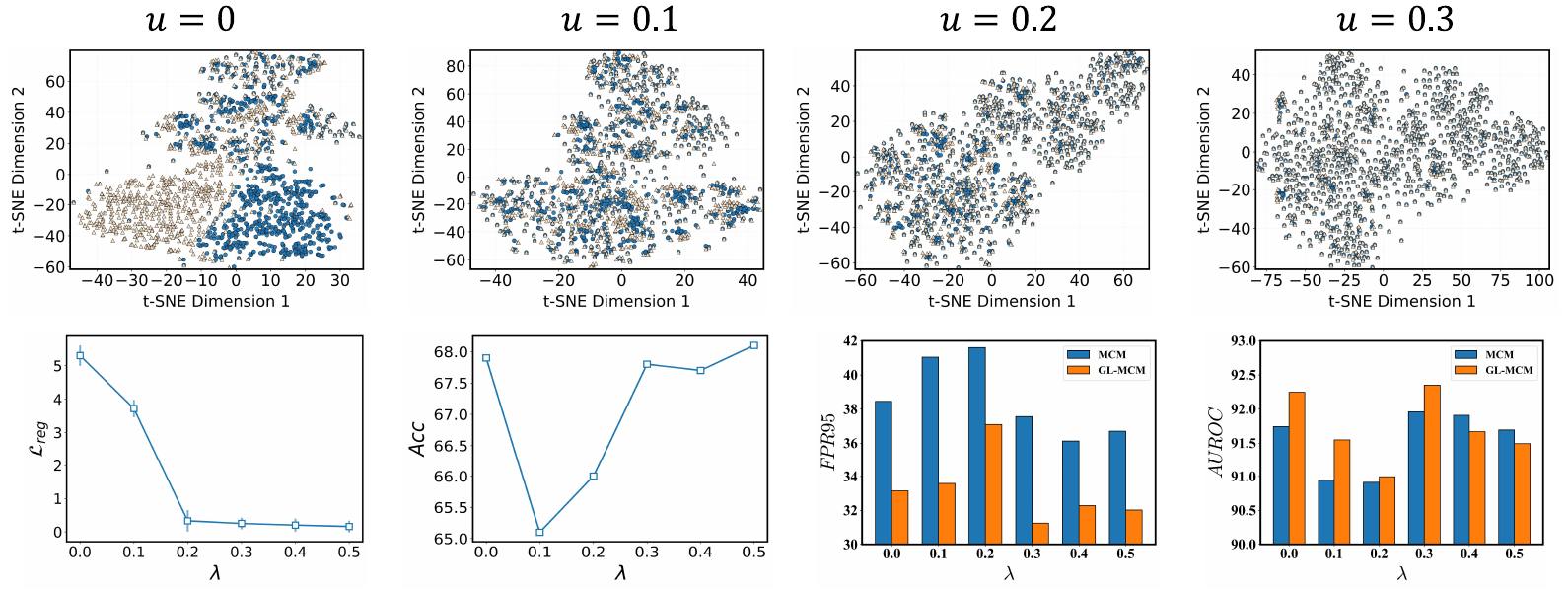}
  \caption{
    Visualization of latent text space representations and model performance landscapes under varying parameter configurations,  demonstrating the alignment between specific trained embeddings (red) and general knowledge distributions (blue),  with superimposed performance contours illustrating the rationality of the GS-balance optimization.}
    \label{fig3}
\end{figure}

\textbf{The discussion of generalization errors.}
The generalization error bound defined in our framework is governed by two loss components. However we we observe that there is a partial knowledge conflict between these two losses, as shown in the Fig. \ref{fig3}, there is a balance between our losses, so this is what we call the GS- balance. The derived bound demonstrates an correlation between the magnitude of our proposed loss terms and the tightness of the generalization guarantee. We conduct theoretical analysis in \ref{sec:A} to elucidate the relationship between the loss and generalization performance. Our empirical results demonstrate that both the training loss and regularization loss exhibit positive correlations with model performance across diverse OOD datasets, thereby validating the theoretical validity of our framework's design principles.

\setlength{\belowcaptionskip}{-5pt}

\section{Conclusion}
This paper addresses the critical challenge of insufficient generalization in few-shot OOD detection by proposing the Generalized Few-shot OOD Detection (\texttt{GOOD}) framework. Our theoretical analysis reveals that preserving the \textit{Generality}-\textit{Specificity} Balance (\textit{GS}-Balance) through general knowledge guidance significantly reduces the upper bound of generalization error. Building on this foundation, we introduced an auxiliary General Knowledge Model (GKM) and developed the Knowledge Dynamic Embedding (KDE) mechanism to dynamically align the OOD detection model with the GKM's outputs. The proposed Generalized Belief (G-Belief) further ensures reliable knowledge transfer by prioritizing high-confidence guidance from the GKM. Extensive experiments across diverse benchmarks demonstrate that \texttt{GOOD} achieves state-of-the-art performance in both conventional and adversarial OOD scenarios. Future work will explore extending this framework to multi-modal settings and enhancing the GKM's adaptability for dynamic open-world environments. The integration of uncertainty quantification mechanisms to refine G-Belief estimation also presents a promising research direction.
\appendix
\renewcommand{\thesection}{Appendix \Alph{section}} 

\section{Proof of Theorem 1}  
\label{sec:A}
Following the setup of \cite{zhang2024best}, we formally define the Total Variation Distance and Disparity Discrepancy as follows:
\newtheorem{definition}{Definition}
\begin{definition}[Disparity with Total Variation Distance]
Given two hypotheses $ f', f \in \mathcal{F} $. Let $D$ be a dataset serving as a collection of instances for variable $x$. we define the Disparity with Total Variation Distance between them as:
\begin{equation}
\label{eq:1}
\mathrm{TV}_{\mathcal{D}}(f, f') = \frac{1}{2} \mathbb{E}_{\mathcal{D}} \left[ \sum_{i=1}^C \left| f_i(x) - f'_i(x) \right| \right],
\end{equation}
\end{definition}

\begin{definition}[Disparity Discrepancy with Total Variation Distance]
Given a hypothesis space $ \mathcal{F} $ and two distributions $ P, Q $, the Disparity Discrepancy with Total Variation Distance is defined as:
\begin{equation}
    d_{\mathcal{F}}(P, Q) := \sup_{f', f \in \mathcal{F}} \left( \mathrm{TV}_P(f', f) - \mathrm{TV}_Q(f', f) \right)
\end{equation}
\end{definition}

For any three hypothesis distributions $f^{1}$, $f^{2}$ and $f^{3}$, the Total Variation Distance satisfies triangle equality, we have

\begin{equation}
\begin{aligned}
 & \mathrm{TV}_D(f^{1},f^{2})\leq\mathrm{TV}_D(f^{1}, f^{3})+\mathrm{TV}_D(f^{2}, f^{3}), \\
 & \mathrm{TV}_D(f^{1},f^{2})\geq\mathrm{TV}_D(f^{1}, f^{3})-\mathrm{TV}_D(f^{2}, f^{3}).
\end{aligned}
\end{equation}

Following Yao\cite{yao2024out}, the generalization error of an OOD detector $f$ is defined as:
\begin{equation}
\label{eq:2}
\mathrm{GError}(f) = \mathrm{TV}_{\mathcal{D}}(f, \mathcal{P}_{\text{true}}).
\end{equation}
Where our groundtruth distribution $\mathcal{P}_{true}$ corresponds to the true labels for ID samples $x_{id}$ or a uniform distribution $\mathcal{U}$ over $C$ classes for OOD samples $x_{ood}$.

Let $f^{*}$ be the hypothesis which jointly minimizes the total variance distance between the predicted distribution $f^{*}$ with uniform distribution U for OOD data or groundtruth for ID data , which is to say
\begin{equation}
f^{*}=\underset{f\in\mathcal{F}}{\operatorname*{\operatorname*{argmin}}}\{\mathrm{TV}_{\mathcal{D}_{train}}(f,\mathcal{P}_{true})+\mathrm{TV}_{\mathcal{D}_{test}}(f,\mathcal{P}_{true})\}.
\end{equation}
Set $\lambda = \mathrm{TV}_{\mathcal{D}_{train}}(f^{*},\mathcal{P}_{true})+\mathrm{TV}_{\mathcal{D}_{test}}(f^{*},\mathcal{P}_{true})$, then by the triangle equality we have

\begin{equation}
\begin{aligned}
GError(f)&=\mathrm{TV}_{\mathcal{D}_{test}}(f,\mathcal{P}_{true}) \\
&\leq\mathrm{TV}_{\mathcal{D}_{test}}(f,f^*)+\mathrm{TV}_{\mathcal{D}_{test}}(f^*,\mathcal{P}_{true}) \\
&=\mathrm{TV}_{\mathcal{D}_{test}}(f,f^*)+\mathrm{TV}_{\mathcal{D}_{test}}(f^*,\mathcal{P}_{true})+\mathrm{TV}_{\mathcal{D}_{train}}(f,f^*)-\mathrm{TV}_{\mathcal{D}_{train}}(f,f^*) \\
&=\mathrm{TV}_{\mathcal{D}_{train}}(f,f^*)+\mathrm{TV}_{\mathcal{D}_{test}}(f^*,\mathcal{P}_{true})+\mathrm{TV}_{\mathcal{D}_{test}}(f,f^*)-\mathrm{TV}_{\mathcal{D}_{train}}(f,f^*) \\
&\leq  \mathrm{TV}_{\mathcal{D}_{train}}(f,\mathcal{P}_{true})+\mathrm{TV}_{\mathcal{D}_{train}}(f^{*},\mathcal{P}_{true}) \\
& \quad +\mathrm{TV}_{\mathcal{D}_{test}}(f^{*},\mathcal{P}_{true})+\mathrm{TV}_{\mathcal{D}_{test}}(f,f^{*})-\mathrm{TV}_{\mathcal{D}_{train}}(f,f^{*}) \\
&=\mathrm{TV}_{\mathcal{D}_{train}}(f,\mathcal{P}_{true})+\mathrm{TV}_{\mathcal{D}_{test}}(f,f^{*})-\mathrm{TV}_{\mathcal{D}_{train}}(f,f^{*}) +\lambda
\end{aligned}
\end{equation}
We have a GKM denfined as $g \in \mathcal{F}$
\begin{equation}
\begin{aligned}
GError(f)&\leq \mathrm{TV}_{\mathcal{D}_{train}}(f,\mathcal{P}_{true})+\mathrm{TV}_{\mathcal{D}_{test}}(f^{*},g)+\mathrm{TV}_{\mathcal{D}_{test}}(g,f) \\
& \quad - \mathrm{TV}_{\mathcal{D}_{train}}(f^{*},g) +\mathrm{TV}_{\mathcal{D}_{train}}(g,f)+\lambda \\
& = \mathrm{TV}_{\mathcal{D}_{train}}(f,\mathcal{P}_{true})+\mathrm{TV}_{\mathcal{D}_{test}}(f^{*},g)+\mathrm{TV}_{\mathcal{D}_{test}}(g,f) \\
& = \mathrm{TV}_{\mathcal{D}_{train}}(f,\mathcal{P}_{true}) +\mathrm{TV}_{\mathcal{D}_{train}}(g,f) +\mathrm{TV}_{\mathcal{D}_{test}}(g,f) \\
& \quad +\mathrm{TV}_{\mathcal{D}_{test}}(f^{*},g)  -\mathrm{TV}_{\mathcal{D}_{train}}(f^{*},g) +\lambda
\label{eq:15}
\end{aligned}
\end{equation}

Due to the inherent limitations in our training data distribution $\mathcal{D}_{train}$, the discrepancy function $d_{\mathcal{F}}$ fails to achieve the theoretical supremum during empirical risk minimization. Prior work \cite{zhang2024best} hypothesizes that near OOD detection can asymptotically approach zero. In order to circumvent the unachievable condition of supremum, we use a pre-trained foundation model g from large-scale dataset, formally defining the discrepancy function $d_{\mathcal{F}}(\mathcal{D}_{train}, \mathcal{D}_{test})$ as $\mathrm{TV}_{\mathcal{D}_{test}}(f^{*}, g) - \mathrm{TV}_{\mathcal{D}_{train}}(f^{*}, g)$. It is evident that this function has become a constant due to the introduction of GKM, which represents the gap between the training set and the test set as perceived by GKM. Consequently, Eq. \ref{eq:15} is expressed as follows:
\begin{equation}
\begin{aligned}
GError(f)
&\leq \mathrm{TV}_{\mathcal{D}_{train}}(f,\mathcal{P}_{true}) +\mathrm{TV}_{\mathcal{D}}(g,f) +d_{\mathcal{F}}(\mathcal{D}_{train}, \mathcal{D}_{test}) +\lambda
\label{eq:16}
\end{aligned}
\end{equation}

Applying Pinsker's Inequality, the following inequality holds
\begin{equation}
\mathcal{L}_{train}(f) \propto  KL(f||\mathcal{P}_{true})\geq2TV(F_{f}(x)||\mathcal{P}_{true})^{2}
\label{eq:17}
\end{equation}

We provide a recap of the Rademacher complexity measure for modelling complexity. We use a complexity-based learning theory to quantify generalisation error. Let $\mathcal{D}_{train}$ be the training dataset. Firstly, we consider the empirical error of $f$. Then, for $1 > {\delta} > 0$ there is at least a 1 -${\delta}$  probability that

\begin{equation}
\begin{aligned}
\mathrm{TV}_{\mathcal{D}_{train}}(f,\mathcal{P}_{true})
&\leq {\hat{\mathrm{TV}}}_{\mathcal{D}_{train}}(f,\mathcal{P}_{true})+\mathcal{R}_{m}(\mathcal{H})+\sqrt{\frac{ln(\frac{1}{\delta})}{2M}} \\
&\leq\sqrt{\frac{1}{2}\mathrm{KL}_{\mathcal{D}_{train}}(f,\mathcal{P}_{true})}+\mathcal{R}_{m}(\mathcal{H})+\sqrt{\frac{ln(\frac{1}{\delta})}{2M}} 
\label{eq:18}
\end{aligned}
\end{equation}
where ${\hat{\mathrm{TV}}}$ is the empirical error of $f$ and $\mathcal{R}_{m}(\mathcal{H})$ is the Rademacher complexities..

So, We use Eq.  \ref{eq:16} and Eq. \ref{eq:18}. Finally the generalized error is:
\begin{equation}
GError(f) \leq 
\underbrace{\sqrt{\frac{1}{2}\mathrm{KL}_{\mathcal{D}_{\text{train}}}(f,\mathcal{P}_{\text{true}})}}_{\text{Train Loss}} + 
\underbrace{\mathrm{TV}_{\mathcal{D}}(g,f)\vphantom{\sqrt{\frac{1}{2}\mathrm{KL}_{\mathcal{D}_{\text{train}}}}}}_{\text{Regularization Loss}} + 
d_{\mathcal{F}}(\mathcal{D}_{\text{train}}, \mathcal{D}_{\text{test}}) +
C
\label{eq:19}
\end{equation}
where $C = \mathcal{R}_{m}(\mathcal{H})+\sqrt{\frac{ln(\frac{1}{\delta})}{2M}}+\lambda$ represents a constant. 

The first item proportional to our train loss, the second item proportional to our regularization loss. As demonstrated in Eq. \ref{eq:17}, the initial term is proportional to the model's training loss, while the subsequent term necessitates the restriction of the distance between the logit of our model and GKM on the dataset. Therefore, the second term is to be regarded as the regularization term. In order to guarantee that the regularisation loss can function in such a manner as to reduce the discrepancy between the outputs of the training set and the test set, it is generally necessary to ensure that the fine-tuning methods employed are consistent. One such method that may be employed is the prompt-tuning method, which is selected for both the training set and the test set.

\section{OOD Score}  
 The CLIP model's multimodal feature alignment capability enables the MCM \cite{ming2022delving} method to perform zero-shot OOD detection by quantifying the similarity distribution between image features and $C$ class text embeddings. The OOD Score function is defined as follows:

\begin{equation}
    S_{\text{$_{MCM}$}} = \max_{i} \frac{\exp\left(\langle \phi_I(\mathbf{x}), \phi_T(\mathbf{t}_i) \rangle / \tau \right)}{\sum_{j=1}^{C} \exp\left(\langle \phi_I(\mathbf{x}), \phi_T(\mathbf{t}_{j}) \rangle / \tau \right)}
\end{equation}

where $\tau = 1$ is the temperature parameter, and $\langle \cdot, \cdot \rangle$ denotes cosine similarity.

By introducing a global-local hierarchical feature matching mechanism, GL-MCM \cite{miyai2025gl} extends the OOD score calculation to:
\begin{equation}
S_{\text{$_{GL-MCM}$}} = \max_{i} \frac{\exp\left(\langle \phi_I(\mathbf{x}^{local}), \phi_T(\mathbf{t}_i) \rangle / \tau \right)}{\sum_{j=1}^{C} \exp\left(\langle \phi_I(\mathbf{x}^{local}), \phi_T(\mathbf{t}_i) \rangle / \tau \right)} + S_{\text{$_{MCM}$}} 
\label{eq:glmcm}
\end{equation}
where $\mathbf{x}^{local}$ represents the feature of the $i$-th local image patch.

\section{Experimental Details}  
\textbf{Base OOD Benchbark}. The implementation of the system adheres to the LoCoOp framework with CLIP-ViT-B/16 \cite{Dosovitskiy_Beyer_Kolesnikov_Weissenborn_Zhai_Unterthiner_Dehghani_Minderer_Heigold_Gelly_et}, where the feature maps exhibit a spatial resolution of 14x14. The key hyperparameters have been empirically configured as follows: the neighbourhood size K = 200 across all experiments, the knowledge distillation coefficient $\alpha$ = 0.25, and the regularization weight $\lambda$ = 0.3. The additional training specifications encompass 50 epochs with a base learning rate of 0.002, a batch size of 32, and a prompt token length of N=16. It is imperative that all experiments are conducted on a single NVIDIA A6000 GPU in order to ensure hardware consistency.

\textbf{Hard OOD Benchbark}. It is evident that our fundamental experimental details are consistent with those of the baseood benchmark. However, given that imagenet-10 and imagenet-20 contain 10 and 20 data types respectively, it was determined that the neighborhood size K=2 would be employed for these hard-to-imitate experiments. The results of the model under the 16-shot setting are presented in full in our paper.

\textbf{OpenOOD OOD Benchbark}. The experimental details are fundamentally analogous to the base food benchmark. The imagenet1k has been selected as the ID dataset, while the SSh-hard, NINCO and OpenImage-O have been designated as the OOD dataset. It should be noted that iNaturalist and Texture have not been included in the evaluation process, as these two datasets have previously been evaluated in the base OOD benchmark.

\section{The Selection of A Suitable General Knowledge Model}  
\label{sec:D}
The following experiments are presented, in which other models of general knowledge are selected to guide the model in acquiring general knowledge. The POMP paper \cite{ren2023prompt} was selected as the secondary general knowledge model to present the experimental results. POMP presented the results of prompt tuning on the ImageNet-21K dataset. In this instance, the model under discussion was employed. It is evident that the parameter settings are consistent with the base OOD benchmark.  Our results are shown in Table \ref{result_1k_backbone}, where the clip subscript represents our general knowledge as " a photo of ", and the POMP subscript represents this general knowledge after training on Imagenet-21k. Our results demonstrate that different GKM models can exhibit varying performance for our method, indicating that our model will acquire distinct general knowledge under distinct GKM settings.
\begin{table*}[h]
\caption{The cross-domain generalisation performance of prompt-tuned general knowledge models, pre-trained on ImageNet-21K and evaluated through out-of-distribution benchmarks.}
\label{result_1k_backbone}
\renewcommand{\arraystretch}{1.05}
\centering
\renewcommand\arraystretch{1.1}
\resizebox{1.0\linewidth}{!}{
\begin{tabular}{ccccccccccc}
\hline
\multirow{3}*{Method} &\multicolumn{10}{c}{OOD Dataset} \\
~  & \multicolumn{2}{c}{iNaturalist} & \multicolumn{2}{c}{SUN} & \multicolumn{2}{c}{Places} & \multicolumn{2}{c}{Texture} & \multicolumn{2}{c}{Average}\\
\cmidrule(lr){2-3} \cmidrule(lr){4-5} \cmidrule(lr){6-7}  \cmidrule(lr){8-9} \cmidrule(lr){10-11} 
~  &FPR95	&AUROC  &FPR95	&AUROC  &FPR95	&AUROC  &FPR95	&AUROC  &FPR95	&AUROC\\
\hline


& \multicolumn{10}{c}{MCM}\\
\texttt{GOOD}$_\texttt{CLIP}$          & $\textbf{27.74}$ & $94.16$ & $34.78$ & $93.01$ & $42.55$ & $90.19$ & $48.48$ & $89.05$ & $38.39$ & $91.60$ \\
\texttt{\texttt{GOOD}}$_\texttt{POMP}$      &$30.80$  &$\textbf{94.17}$ &$\textbf{31.25}$ 	&$\textbf{93.91}$	&$\textbf{39.78}$ 	&$\textbf{90.79}$	&$\textbf{41.50}$ 	&$\textbf{90.81}$ &$\textbf{35.83}$ 	&$\textbf{92.42}$ \\
\hline
& \multicolumn{10}{c}{GL-MCM}\\
\texttt{GOOD}$_\texttt{CLIP}$       & $\textbf{13.59}$ & $\textbf{96.81}$ & $27.73$ & $93.87$ & $35.94$ & $91.09$ & $51.21$ & $85.80$ & $32.12$ & $91.89$ \\
\texttt{GOOD}$_\texttt{POMP}$      &$16.41$ 	&$96.48$	&$\textbf{22.78}$ 	&$\textbf{95.05}$	&$\textbf{32.41}$ 	&$\textbf{91.80}$	&$\textbf{44.11}$ 	&$\textbf{88.95}$  &$\textbf{28.92}$ 	&$\textbf{93.07}$ \\

\hline
\end{tabular}
}
\vspace{-5pt} 
\end{table*}

Moreover, in order to demonstrate the rationality of our methodology, we employ the same comparison strategy as outlined in Table 1.  The results of the ood score of POMP using MCM and GL-MCM in ood detection are presented, as well as the results of the ood score of the LoCoOp model using only our training loss.  The following presentation will outline the output results of the model under the KDE strategy.  The results of the study are presented in tabular form.  The findings of this study suggest that the proposed methodology explores the upper limit of OOD detection, while exhibiting the POMP generalization.
\setlength{\belowcaptionskip}{-3pt}
\begin{table*}[h]
\caption{The model performance of POMP when used as the GKM model. The present method has been developed in such a manner that it inherits the generalisation ability of POMP, whilst also exploring the upper limit of OOD detection.}
\label{result_1k_backbone_abalition}
\renewcommand{\arraystretch}{1.05}
\centering
\renewcommand\arraystretch{1.1}
\resizebox{1.0\linewidth}{!}{
\begin{tabular}{ccccccccccc}
\hline
\multirow{3}*{Method} &\multicolumn{10}{c}{OOD Dataset} \\
~  & \multicolumn{2}{c}{iNaturalist} & \multicolumn{2}{c}{SUN} & \multicolumn{2}{c}{Places} & \multicolumn{2}{c}{Texture} & \multicolumn{2}{c}{Average}\\
\cmidrule(lr){2-3} \cmidrule(lr){4-5} \cmidrule(lr){6-7}  \cmidrule(lr){8-9} \cmidrule(lr){10-11} 
~  &FPR95	&AUROC  &FPR95	&AUROC  &FPR95	&AUROC  &FPR95	&AUROC  &FPR95	&AUROC\\
\hline
& \multicolumn{10}{c}{MCM}\\
POMP      &$39.65$ 	&$92.86$	&$32.44$ 	&$93.11$	&$39.30$ 	&$90.73$	&$\textbf{40.97}$ 	&$90.34$	&$38.09$ 	&$91.76$ \\
LoCoOp      &$38.96$  &$92.34$ &$32.40$ 	&$93.60$	&$37.95$ 	&$\textbf{91.00}$	&$49.32$ 	&$88.70$ &$39.65$ 	&$91.41$ \\
\rowcolor{gray!40}
\texttt{\texttt{GOOD}}     &$\textbf{30.80}$  &$\textbf{94.17}$ &$\textbf{31.25}$ 	&$\textbf{93.91}$	&$\textbf{39.78}$ 	&$90.79$	&$41.50$ 	&$\textbf{90.81}$ &$\textbf{35.83}$ 	&$\textbf{92.42}$ \\
& \multicolumn{10}{c}{GL-MCM}\\
POMP       &$16.51$ 	&$\textbf{96.53}$	&$26.12$ 	&$93.47$	&$33.48$ 	 &$91.14$   &$44.52$    &$86.93$	&$30.15$ 	&$92.02$ \\
LoCoOp      &$24.38$ 	&$94.95$	&$25.45$ 	&$94.77$	&$32.63$ 	&$\textbf{91.81}$	&$52.32$ 	&$86.58$  &$33.69$ 	&$92.03$ \\
\rowcolor{gray!40}
\texttt{GOOD}      &$\textbf{16.41}$ 	&$96.48$	&$\textbf{22.78}$ 	&$\textbf{95.05}$	&$\textbf{32.41}$ 	&$91.80$	&$\textbf{44.11}$ 	&$\textbf{88.95}$  &$\textbf{28.92}$ 	&$\textbf{93.07}$ \\

\hline
\end{tabular}
\vspace {-5.0em}}
\end{table*}

\section{More Experimental Results}  
The appendices to this section contain further experimental results of our model, the purpose of which is to demonstrate its experimental performance. The following presentation comprises the experimental results of MCM and GL-MCM under a variety of conditions.

The experimental results obtained under the OpenOOD and MCM benchmarks demonstrate that GL-MCM exhibits superior performance in cross-dataset ID and OOD detection scenarios when compared to the baseline MCM. However, when evaluated within the same dataset containing different class partitions, the method exhibits inconsistent performance, with alternating advantages and disadvantages. It is hypothesised that this discrepancy may stem from the higher similarity of local features between ID and OOD samples originating from the same dataset. The proximity of such inherent features has the potential to diminish the discriminative power of local information, thereby hindering effective ID and OOD separation.
\setlength{\abovecaptionskip}{2pt}
\setlength{\belowcaptionskip}{-3pt}
\begin{table*}[h]
\centering
\caption{OOD detection performance for ImageNet-1k as ID, the SSh-hard, NINCO, OpenImage-O as OOD dataset.}
\label{tab:ood_results}
\setlength{\tabcolsep}{4pt} 
\small
\begin{tabular}{@{}lcccccc@{}}
\toprule
\multirow{2}{*}{Method} & 
\multicolumn{6}{c}{\textbf{ImageNet-1K}} \\
\cmidrule(lr){2-5} \cmidrule(l){6-7}
& \multicolumn{2}{c}{SSh-hard} & \multicolumn{2}{c}{NINCO} & 
\multicolumn{2}{c}{OpenImage-O} \\
\cmidrule(lr){2-3} \cmidrule(lr){4-5} \cmidrule(l){6-7}
& FPR95 & AUROC& FPR95 & AUROC& FPR95 & AUROC\\
\midrule
$\texttt{GOOD}_\texttt{{MCM}}$ & $88.78$  & $64.41$ & $79.19$  & $74.10$ & $41.43$  & $91.84$ \\
$\texttt{GOOD}_\texttt{{GL}}$     & $85.13$  & $68.27$ & $72.57$  & $76.06$ & $34.59$  & $92.36$ \\
\bottomrule
\end{tabular}
\vspace{-5pt} 
\end{table*}

The experimental findings yielded from the execution of MCM benchmarks demonstrate that GL-MCM evinces superior performance in OOD detection scenarios when contrasted with the baseline MCM. This outcome is congruent with our experimental expectations and concomitantly signifies that GL-MCM also attains comparatively favourable enhancement results for GL-MCM of GKM.
\setlength{\abovecaptionskip}{2pt}
\setlength{\belowcaptionskip}{-1pt}
\begin{table*}[h]
\centering
\caption{OOD detection performance for ImageNet-10, ImageNet-20 as ID, the corresponding imagenet20, imagenet10 as ood datasetas.}
\begin{tabular}{@{}lccccc@{}}
\toprule
\multirow{2}{*}{Method} & \multicolumn{2}{c}{ImageNet10}  & \multicolumn{2}{c}{ImageNet20} \\
 & \multicolumn{2}{c}{ImageNet20}  & \multicolumn{2}{c}{ImageNet10}  \\
\cmidrule(lr){2-3}\cmidrule(lr){4-5}
 &FPR95&AUROC &FPR95 &AUROC  \\
\midrule
$\texttt{GOOD}_\texttt{{MCM}}$  & $5.70$ & $98.60$ & $16.10$ & $97.66$ \\
$\texttt{GOOD}_\texttt{{GL}}$  & $10.60$ & $98.66$ & $9.90$ & $98.32$ \\
\bottomrule
\end{tabular}
\vspace{-5pt} 
\end{table*}

The subsequent presentation will expound upon the findings of the model's image detection process in relation to imaget100, which will be utilised as the ID data.  The experimental results of the model on 4-shot are also presented.  In the present experiment, the value of K was set to 20.  The 1-shot configuration was not selected as the experimental outcome due to the inability of our model to converge on the original LoCoOp setting. In order to conduct a one-shot experiment, it is necessary to enlarge the epoch under the LoCoOp setting until the experimental results obtained are consistent with those reported in the aforementioned paper.  The present study employs imagenet-100 as the ID dataset, thereby adopting a methodology that explores enhanced object detection while ensuring the efficacy of the GKM model.   This approach is employed to demonstrate the efficacy of the proposed methodology.
\begin{table*}[h]
\caption{Cross-domain generalization performance on ImageNet-100 as ID data under four-shot learning protocol. A comparison was made between MCM and LoCoOp.}
\label{result_1k_imagenet100}
\renewcommand{\arraystretch}{1.05}
\centering
\renewcommand\arraystretch{1.1}
\resizebox{1.0\linewidth}{!}{
\begin{tabular}{ccccccccccc}
\hline
\multirow{3}*{Method} &\multicolumn{10}{c}{OOD Dataset} \\
~  & \multicolumn{2}{c}{iNaturalist} & \multicolumn{2}{c}{SUN} & \multicolumn{2}{c}{Places} & \multicolumn{2}{c}{Texture} & \multicolumn{2}{c}{Average}\\
\cmidrule(lr){2-3} \cmidrule(lr){4-5} \cmidrule(lr){6-7}  \cmidrule(lr){8-9} \cmidrule(lr){10-11} 
~  &FPR95	&AUROC  &FPR95	&AUROC  &FPR95	&AUROC  &FPR95	&AUROC  &FPR95	&AUROC\\
\hline
&  \multicolumn{10}{c}{MCM}\\
MCM     &$12.58$ 	&$97.27$	&$32.50$ 	&$94.93$	&$30.72$ 	&$94.86$	&$38.33$  &$93.21$	&$28.53$ 	&$95.07$ \\
LoCoOp$_\texttt{MCM}$ &$18.69$  &$96.54$ &$21.16$  &$96.32$ &$27.82$  &$95.12$ &$26.17$  &$94.99$ &$23.46$  &$95.74$ \\
\rowcolor{gray!40}
\texttt{GOOD}$_\texttt{MCM}$ &$\textbf{10.70}$  &$\textbf{97.71}$ &$\textbf{16.81}$  &$\textbf{96.92}$ &$\textbf{22.52}$  &$\textbf{95.65}$ &$\textbf{24.68}$  &$\textbf{95.49}$ &$\textbf{18.67}$  &$\textbf{96.44}$ \\
\hline
& \multicolumn{10}{c}{GL-MCM}\\
GL-MCM      &$\textbf{3.17}$ 	&$\textbf{98.92}$	&$17.94$ 	&$96.85$	&$19.83$ 	 &$\textbf{96.29}$   &$36.20$   &$92.47$	&$19.28$ 	&$96.13$ \\
LoCoOp$_\texttt{GL}$  &$12.97$  &$97.09$ &$\textbf{12.55}$  &$97.20$ &$\textbf{18.15}$  &$96.06$ &$\textbf{26.17}$  &$94.36$ &$17.46$  &$96.18$ \\
\rowcolor{gray!40}
\texttt{GOOD}$_\texttt{GL}$  &$4.44$  &$98.87$ &$13.15$  &$\textbf{97.42}$ &$18.43$  &$96.11$ &$27.23$  &$\textbf{94.48}$ &$\textbf{15.81}$  &$\textbf{96.72}$ \\

\hline
\end{tabular}
}
\vspace{-5pt} 
\end{table*}

\section{GS-balance Analysis}

\begin{wrapfigure}{rh}{4.5cm} 
\includegraphics[width=0.33\textwidth]{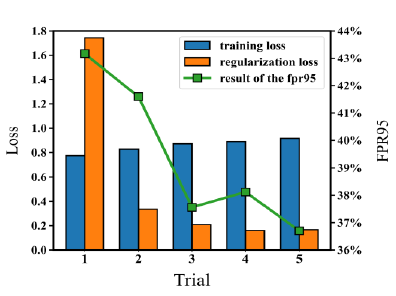}\
\caption{
    The performance of the training loss and regularisation loss in relation to the OOD detection metric.}
\label{fig5}
\vspace{-5pt} 
\end{wrapfigure}
In this section, the relationship between training loss, regularization loss and OOD detection performance is investigated through the lens of GS-balance.  In the course of a series of controlled experiments in which regularization coefficients $\lambda$ were varied, a consistent positive correlation was observed between the total loss values of the models and the OOD detection metrics when the models were sorted by total loss in an ascending order.  Moreover, the findings of this study indicate the existence of an inherent discordance between the objectives of optimisation.  As demonstrated in Fig. \ref{fig5}, there is a concurrence between the diminution of regularisation loss and the augmentation of training loss.  This phenomenon is accompanied by an inverse correlation between these two components.  This phenomenon suggests the existence of conflicting knowledge that may hinder concurrent optimisation of both objectives.   It is imperative that optimal GS-balance points are identified, at which both losses maintain a mutually beneficial dynamic.  The experimental results obtained demonstrate that it is imperative to achieve proper GS-balance through this approach in order to obtain optimal OOD detection performance while maintaining model generalisation capabilities.
\section{Dataset}
\subsection{ID Datasets.}
\textbf{ImageNet-1K.} The present study employs the ImageNet-1K benchmark dataset as the ID data source, procured via the official platform (https://www.image-net.org/).  The experimental design of the present study is aligned with the standardised few-shot evaluation protocol that has been established in CLIP, CoOp and LoCoOp. Model validation is conducted using the ImageNet validation set in its canonical form, which comprises 50,000 annotated images that span 1,000 distinct categories.  This configuration is consistent with the evaluation criteria documented in prior research.

\subsection{OOD Datasets.}  
\textbf{iNaturalist.}
The dataset under consideration is comprised of 859,000 biological specimens, which are divided into more than 5,000 taxonomic categories.  The primary focus of the dataset is flora and fauna biodiversity.  In accordance with the established protocol, the evaluation process is conducted using a sample of 10,000 images, selected at random from a total of 110 classes, with the exclusion of those that are already present in the ImageNet-1K database.

\textbf{SUN.}
The scene recognition corpus under consideration contains 130,000 visual instances, which are divided into 397 environmental categories. For the purpose of comparative analysis, a curated subset of 10,000 images has been employed, sampled from 50 ImageNet-disjoint classes.

\textbf{Places.}
Places provides complementary coverage of environmental semantics, mirroring SUN's conceptual scope in scene understanding. The assessment utilises 10,000 images from 50 non-overlapping classes.

\textbf{TEXTURE.}
The present corpus is one that has been specifically compiled for the purpose of this study. It consists of 5,640 high-resolution texture patterns that have been organised into 47 material categories. A comprehensive evaluation is performed using the full dataset.

\textbf{OpenImage-O.}
This rigorously curated visual recognition benchmark comprises 17,632 images that have been manually filtered through multi-stage quality assurance protocols, achieving 7.8× greater scale diversity than ImageNet-O through pixel-coverage optimisation.  

\textbf{SSB-hard.}
Derived from ImageNet-21K's hierarchical ontology through semantic scarcity sampling, this 49,000-image benchmark spans 980 visually complex categories characterised by high inter-class ambiguity.

\textbf{NINCO.}
The dataset contains 5,879 meticulously annotated samples across 64 novel categories, thereby introducing conceptual novelty through systematic exclusion of ImageNet-1K semantic overlaps.

\textbf{ImageNet-10.}
The creation of ImageNet-10 was driven by the necessity to emulate the class distribution of CIFAR-10, while incorporating high-resolution images. The following categories are contained within the dataset, along with their respective class identifiers: The following subject headings have been identified: The following terms are listed: 'warplane' (n04552348), 'sports car' (n04285008), 'brambling bird' (n01530575), 'Siamese cat' (n02123597), 'antelope' (n02422699). The following have been identified: 'Swiss mountain dog' (n02107574), 'bull frog' (n01641577), 'garbage truck' (n03417042), 'horse' (n02389026), and 'container ship' (n03095699).

\textbf{ImageNet-20.}
In order to facilitate the evaluation of hard OODs with realistic datasets, ImageNet-20 has been curated. The dataset under consideration consists of 20 classes that are semantically similar to ImageNet-10. The categories are selected based on the distance in the WordNet synsets. The following categories are contained therein: The following items are listed herewith: The following objects are documented: a sailboat (n04147183), a canoe (n02951358), a balloon (n02782093), a tank (n04389033), a missile (n03773504), and a bullet train (n02917067). The following species were documented: A starfish (n02317335), a spotted salamander (n01632458), a common newt (n01630670), a zebra (n01631663), and a frilled lizard (n02391049). For the purposes of this study, the following taxa were selected: the green lizard (n01693334), the African crocodile (n01697457), the Arctic fox (n02120079), the timber wolf (n02114367), the brown bear (n02132136), the moped (n03785016), the steam locomotive (n04310018), the space shuttle (n04266014) and the snowmobile (n04252077).

\section{Limitations and Broader Impacts}
Our framework requires strict consistency in adaptation methodologies between the target model and the GKM.  Specifically, both components must employ identical parameter-efficient tuning strategies.  Such as the concurrent prompt-tuning of backbone architectures to maintain logit-level alignment.  Disparate adaptation approaches such as combining prompt-tuning with conventional backbone fine-tuning disrupt the regularization mechanism's capacity to minimize train-test output discrepancies, thereby compromising the model's ability to learn transferable knowledge representations. Moreover, the method under discussion merely provides a means of identifying GS-balance. In the course of the process under discussion, it is possible that more efficacious methods may be discovered for the purpose of ascertaining GS-balance. In conclusion, the methodology employed is founded upon the premise that GKM is a model that demonstrates effective generalisation. The upper limit of the detection of ood explored is constrained by the general knowledge of GKM. The capacity for exploration in the context of prompt-tuning is constrained by this limitation. Alternatively, it may be necessary to compromise certain aspects ood detection capabilities in order to facilitate the generalisation of the model.

\bibliography{ref}
\bibliographystyle{iclr2025_conference}
\newpage
\end{document}